\documentclass[10pt,twocolumn,letterpaper]{article}

\usepackage{iccv}
\usepackage{times}
\usepackage{epsfig}
\usepackage{graphicx}
\usepackage{amsmath}
\usepackage{amssymb}
\usepackage{multirow}
\usepackage[hyphens]{url}

\usepackage[breaklinks=true,bookmarks=false]{hyperref}
\usepackage[hyphenbreaks]{breakurl}

\iccvfinalcopy 

\def\iccvPaperID{****} 

\ificcvfinal\fi

\begin{document}

\title{CheXFusion: Effective Fusion of Multi-View Features using Transformers for Long-Tailed Chest X-Ray Classification}

\author{Dongkyun Kim\\
Carnegie Mellon University\\
Pittsburgh, PA 15213, USA\\
{\tt\small dongkyuk@andrew.cmu.edu}
}

\maketitle
\ificcvfinal\thispagestyle{empty}\fi

\begin{abstract}
   Medical image classification poses unique challenges due to the long-tailed distribution of diseases, the co-occurrence of diagnostic findings, and the multiple views available for each study or patient. This paper introduces our solution to the ICCV CVAMD 2023 Shared Task on CXR-LT: Multi-Label Long-Tailed Classification on Chest X-Rays. Our approach introduces CheXFusion, a transformer-based fusion module incorporating multi-view images. The fusion module, guided by self-attention and cross-attention mechanisms, efficiently aggregates multi-view features while considering label co-occurrence. Furthermore, we explore data balancing and self-training methods to optimize the model's performance. Our solution achieves state-of-the-art results with 0.372 mAP in the MIMIC-CXR test set, securing 1st place in the competition. Our success in the task underscores the significance of considering multi-view settings, class imbalance, and label co-occurrence in medical image classification. Public code is available at \url{https://github.com/dongkyuk/CXR-LT-public-solution}.
\end{abstract}

\begin{figure}[t]
\begin{center}
\includegraphics[width=0.9\linewidth]{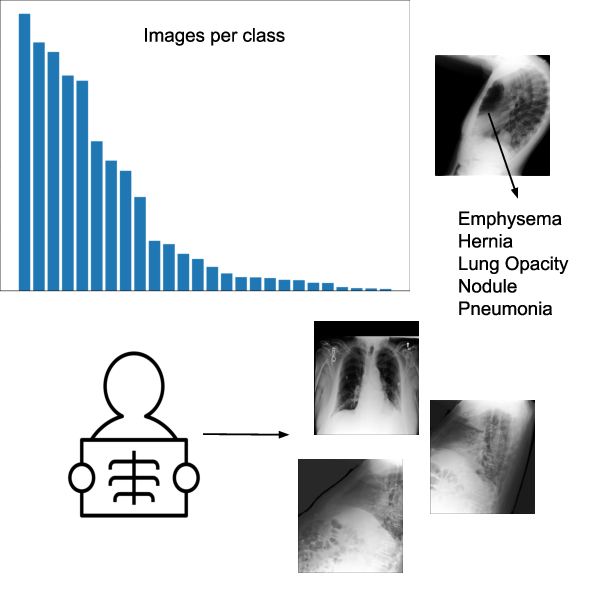}
\end{center}
   \caption{Challenges in medical image classification. Addressing long-tailed distributions, label co-occurrence, and multiple views is crucial for accurate and comprehensive disease diagnosis.}
\label{fig:long}
\label{fig:onecol}
\end{figure} 

\section{Introduction}

The field of medical image classification has gained significant attention due to the increasing recognition of the potential of artificial intelligence in healthcare. However, compared to most 
image classification benchmarks~\cite{imgnet} for standard deep learning methods, medical image classification faces several specific challenges that necessitate novel approaches: 1) Long-tailed distributions, 2) Label co-occurrence, and 3) Multiple views.

Medical image classification exhibits long-tailed distributions, where a small subset of diseases or conditions is commonly observed, while the majority of diseases are relatively rare~\cite{Zhou_2021}. This imbalanced distribution poses a challenge for traditional deep learning methods, as they tend to prioritize the most common classes at the expense of the less frequent ones. Consequently, the accurate identification and diagnosis of rare diseases become challenging, potentially leading to missed or delayed diagnoses.

Medical diagnoses also often involve the presence of multiple diseases, resulting in label co-occurrence. For instance, a patient's chest X-ray may concurrently exhibit findings related to cardiomegaly, consolidation, and edema. Multi-label classification faces several challenges due to label imbalance and the dominance of negative labels~\cite{dbloss}. However, many existing deep learning approaches for medical image classification do not explicitly consider the challenges of deep learning in multi-label settings, which can limit their performance in real-world scenarios.

Finally, medical image classification encompasses multiple views, with different imaging modalities or image perspectives available for each patient or study. For example, chest X-rays can be captured from various angles or with different imaging protocols. Each view provides unique information that can contribute to the accurate diagnosis of diseases. For example, the lateral view shows 15\% of the lung that is unidentifiable from a single posterior anterior view~\cite{RAOOF2012545} and often helps clarify diagnosis \cite{FEIGIN20101560}. However, effectively leveraging multiple views for classification poses additional challenges, requiring appropriate fusion techniques to integrate the information from diverse sources.

The CXR-LT shared task successfully provides a benchmark to address these challenges. It builds upon the MIMIC-CXR-JPG dataset~\cite{johnson2019mimiccxrjpg}, expanding the target classes from 14 to 26 by incorporating 12 new disease findings obtained from radiology reports. This expansion enables a more comprehensive representation of the diagnostic landscape encountered in clinical practice. It also benchmarks images with various numbers of different views, allowing the exploration of multi-view models. 

In this work, we propose CheXFusion, a transformer-based fusion module 
that combines the benefits of self-attention and cross-attention mechanisms to dynamically aggregate multi-view data while accounting for the co-occurrence of different disease findings.

In addition to the fusion module, we employ various data-balancing techniques and self-training strategies to further enhance the performance of our model. These techniques enable better handling of class imbalance, improve the generalization capabilities of the model, and enhance its ability to learn from limited labeled data.

Our solution emerged as the top-performing method in both the validation and test leaderboards of the CXR-LT shared task. Through rigorous experiments, we also verify the effectiveness of our proposed method in addressing the challenges of multi-label, long-tailed medical image classification. By considering the multi-view nature of chest X-rays, class imbalance, and label co-occurrence, our solution contributes to more accurate disease diagnosis, paving the way for improved patient care and treatment outcomes.

Our contribution can be summarized as the following:

\begin{enumerate}
\item We propose CheXFusion, a transformer-based fusion module that effectively integrates features extracted from multi-view medical images, leveraging self-attention and cross-attention mechanisms.
\item We conduct extensive experiments to verify the advantages of incorporating various data balancing techniques and self-training strategies.
\item Our solution achieves top performance in both the validation and test leaderboards of the CXR-LT shared task, demonstrating its effectiveness.
\end{enumerate}

\begin{figure*}
\begin{center}
\includegraphics[width=1\linewidth]
{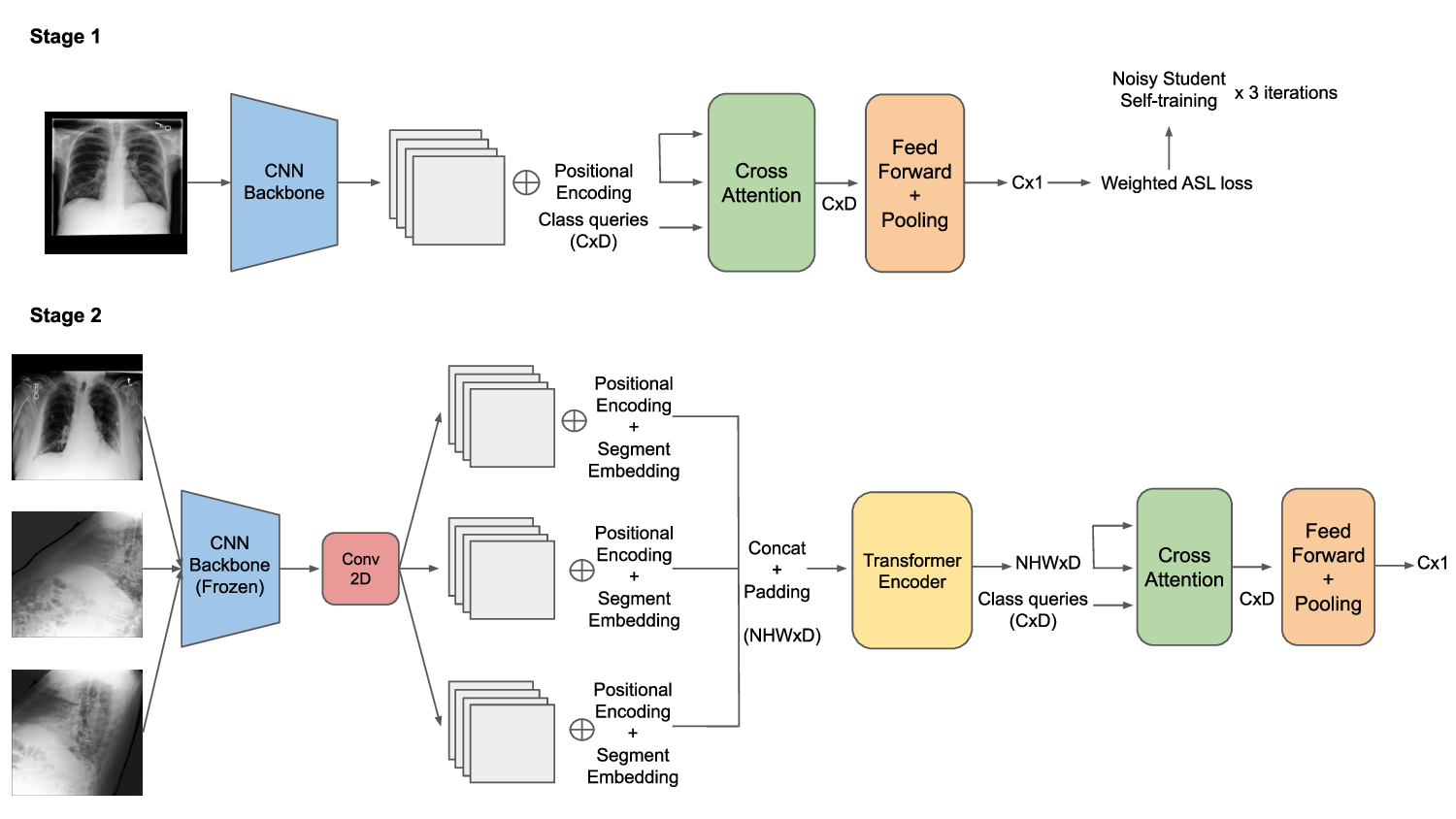}
\end{center}
\centering\caption{The overall pipeline of our proposed solution CheXFusion. $C$ is the number of classes, $D$ is the token dimension, and $N$ is the maximum number of images that can be used per patient.}
\label{fig:short}
\end{figure*}

\section{Related Work}

\subsection{Multi-label Classification}
A simple approach to multi-label classification is to train multiple binary classifiers, where each classifier predicts the presence or absence of a specific label independently~\cite{Multi-Label}. However, such a naive approach fails to consider label co-occurrence and class imbalance, resulting in suboptimal performance.

Early approaches to addressing these problems include transforming the multi-label problem into a multi-class with Label Powerset~\cite{Tsoumakas}, building a chain of binary classifiers~\cite{ClassifierChains} or Multi-Label k-Nearest Neighbors (MLkNN)~\cite{ML-KNN}. 

Recently, many deep learning-based approaches have been proposed with much success. The proposed methods can be categorized
into three categories: improving loss functions to consider class imbalance~\cite{dbloss, asl}, modeling label correlations~\cite{chen2019learning, ye2020attentiondriven} and locating regions of interest~\cite{wang2017multilabel, liu2021query2label, ridnik2021mldecoder}. This paper uses a mixture of these methods with the proposed transformer-fusion model.

\subsection{Multi-modal Learning}
Multi-modal or (multi-view) learning aims to leverage multiple perspectives or sources of information to improve the performance of machine learning models. In medical image classification, multi-view learning becomes essential when different imaging modalities, imaging protocols, or image perspectives are available for each patient or study.

Various methods have been proposed for multi-view learning for chest radiograph classification. Rubin \etal ~\cite{rubin2018large} proposed DualNet, where posterior anterior and lateral views are processed by two separate DenseNet-based CNNs and concatenated before being passed to the classification head. 
Zhu  \etal ~\cite{mvcnet} proposed MVC-Net, where a third BPT branch was utilized for fusing intermediate representations, and a consistency loss was applied between the frontal and lateral branches. These methods all utilize separate, fixed branches for each modality and thus face four significant problems: it \emph{linearly scales to the number of modalities used, assumes all modalities to be present, require knowledge of the modality, and is not applicable for multiple images of the same view}. Kohankhaki \etal ~\cite{10020356} partially addressed these issues with a transformer encoder approach utilizing two alternating cross-attention steps, attending to lateral features based on frontal features and vice versa.

However, the alternating cross-attention approach still requires knowledge of the modality and cannot address cases where multiple images exist for the same view. In addition, it was only tested for the binary classification of pneumonia, which is different from our task. To overcome these limitations, we propose CheXFusion. This transformer-based fusion model does \emph{not require modality information, is invariant to the number of samples, and is easily compatible} with existing solutions to multi-label classification.

\subsection{Transformer-Based Models}

Transformers~\cite{vaswani2017attention} have revolutionized natural language processing tasks by effectively capturing long-range dependencies and modeling sequences. They have also shown promise in computer vision tasks, including image classification and object detection.

Vision Transformers (ViTs)~\cite{dosovitskiy2021image}, have been successful in image classification by representing images as sequences of patches and applying transformer encoders to capture their relationships. Carion \etal proposed DETR~\cite{carion2020endtoend}, a transformer-based
end-to-end object detection framework. Transformers have also been applied to multi-label classification tasks, where they excel at modeling label dependencies and attending to separate regions of interest through attention mechanisms. Liu \etal proposed Query2Label~\cite{liu2021query2label}, where
a transformer decoder architecture was used with label embeddings as queries to probe a vision backbone generated feature map. Ridnik \etal designed Ml-decoder~\cite{ridnik2021mldecoder}, a variant of query2label redesigned for scalability while maintaining similar performance. 

Our proposed transformer-based fusion module builds upon these advancements and extends them to handle the challenges of multi-label, long-tailed medical image classification with multiple views. We utilize a transformer encoder to aggregate multi-view features and utilize Ml-decoder~\cite{ridnik2021mldecoder} as the final cross-attention multi-label classification head.

\section{Method}
\subsection{Overview}

The overall pipeline is depicted in Figure 2. Our proposed method consists of two stages: we first train a single-view convolutional neural network backbone, and in the second stage, we freeze the pre-trained backbone and train a transformer-based fusion module named ChexFusion that encodes multi-view features with minimal complexity and high flexibility.

\subsection{Backbone pre-training}
The first stage of our method involves pre-training a single-view convolutional neural network (CNN) backbone. This backbone serves as the feature extractor for each view in the subsequent fusion stage. The backbone is trained on all views, allowing it to serve as a general feature extractor that can handle diverse views.

A commonly used approach for classification heads in computer vision is to use a variant of global average pooling (GAP) followed by a fully connected layer~\cite{726791}. However, such GAP-based classification heads are not fit for the multi-label classification task, where we need to attend to different characteristics in order to identify
different classes and objects. 

Thus, we employ Ml-Decoder~\cite{ridnik2021mldecoder}, a transformer-based classification head with the removal of redundant self-attention blocks and a group-decoding scheme for class number scalability. In contrast to the original implementation, we did not find grouped queries necessary as the number of classes in our task (26 classes) was much smaller than the number of classes explored in the original paper (1000+ classes). Our experiments show that the change of classification heads alone provides a substantial performance increase.

Though not further explored in this paper, the use of Ml-Decoder may also allow zero-shot learning by utilizing word embeddings as class queries and applying query augmentations during training. 

\subsection{Transformer Fusion Model}
The second stage of our method involves training a transformer-based fusion module named CheXFusion. This module takes as input the features extracted by the pre-trained backbone from multiple views and effectively integrates them to perform multi-label classification.

The architecture of CheXFusion is inspired by the transformer-based models used in computer vision~\cite{carion2020endtoend}. It consists of a transformer encoder that encodes the features of each view, followed by a Ml-decoder~\cite{ridnik2021mldecoder} that performs multi-label classification based on the fused features.

Starting from the initial images $x \in R^{N_0 \times H_0 \times W_0 \times D_0}$, where $N_0$ is the number of available images for a patient, the pre-trained CNN backbone followed by an additional convolutional layer generates a lower-resolution activation map $f \in R^{N_0 \times H \times W \times D}$. The CNN backbone is kept frozen to enable faster training and to allow the fusion module to focus on learning how to integrate the multiple features. By default, we set $D = 768$ and $H, W = 16, 16$.

Since $N_0$ is different for each patient across a batch, we concatenate a learnable padding token to obtain a feature map $f \in R^{N \times H \times W \times D}$, where $N$ is the maximum number of images that can be used per patient. In rare cases when $N_0 > N$, we sampled $N$ images and did not use the rest. By default, we set $N=4$. 

Following DETR~\cite{carion2020endtoend}, we add a 2d sinusoidal positional encoding to each feature map to take account of the permutation-invariance of transformer architectures. Similar to BERT~\cite{devlin2019bert}, we add a different learnable segment embedding to each feature map to indicate which image it belongs to. Since the order of the feature maps is irrelevant to the task, we shuffle them along the first index.
Finally, as the transformer encoder expects a sequence as input, we reshape the features into a sequence of tokens $s \in R^{NHW \times D}$.

We use the standard transformer encoder architecture, with multiple layers comprising a multi-head self-attention mechanism and a feed-forward neural network. In our case, the feature aggregation performed by the transformer encoder is equivalent to encoding multiple sentences in natural language processing, where each sentence corresponds to a single chest X-ray image. 

The decoding phase is identical to stage 1, where we employ a Ml-Decoder~\cite{ridnik2021mldecoder} classification head. Overall, CheXFusion is similar to a text-to-text encoder-decoder transformer, where we encode multiple images by considering each image feature map as a sentence of features and decode it based on multiple class queries~\cite{vaswani2017attention, raffel2020exploring}.

CheXFusion can be used as a simple plug-and-play method for any multi-label classification tasks utilizing multiple views. The design makes it scalable and flexible, with state-of-the-art results.

\subsection{Loss Function}
Multi-label long-tailed classification suffers from two types of imbalance, namely \emph{inter-class} imbalance and \emph{intra-class} imbalance.

Inter-class imbalance occurs from the long-tailed distribution of the dataset. Using a vanilla binary cross-entropy loss function leads to uneven gradient updates for different classes, resulting in an imbalanced training that degrades performance~\cite{zhang2023deep}. To address this, we use a weighted binary cross-entropy loss:
\begin{equation}
L_{wbce} = -\sum_{i=1}^{C} w_i \left(y_i \log(p_i) + (1-y_i) \log(1-p_i)\right)
\end{equation}
where $C$ is the total number of classes, and $y_i$, $p_i$, and $w_i$ are ground truth labels, predicted probability, and weight for class $i$. $w_{i} = y_ie^{1-\rho} + (1-y_i)e^{\rho}$ where $\rho$ is the ratio of positive samples for class $i$.

Intra-class imbalance occurs from the dominance of negative labels in multi-label classification. Binary cross entropy (BCE) loss is symmetric, with uniform weights to positive and negative classes. This leads to an over-suppression of the positive classes, as the model is penalized heavily for false positives. To address this, we utilize asymmetric loss~\cite{asl}, a variant of focal loss.
\begin{equation}
\begin{split}
L_{asl} = -\sum_{i=1}^{C} (1-p_i)^{\gamma_{+}} y_i \log(p_i) + \\ 
p_{mi}^{\gamma_{-}}(1-y_i) \log(1-p_{mi})
\end{split}
\end{equation}
where $p_{mi} = \max(p_i-m,0)$. By default, we set $\gamma_{+} =1$, $\gamma_{-} =4$, $m=0.05$ in our experiments.

By combining the weighted binary cross-entropy loss with asymmetric loss, we can effectively handle both inter-class and intra-class imbalances in the multi-label long-tailed classification task.
\begin{equation}
\begin{split}
L = -\sum_{i=1}^{C} w_i ((1-p_i)^{\gamma_{+}} y_i \log(p_i) + \\
p_{mi}^{\gamma_{-}}(1-y_i) \log(1-p_{mi}))
\end{split}
\end{equation}

\subsection{Self-Training}
Various transfer learning methods have been explored in long-tailed learning to improve performance by introducing additional information and positive samples into model training~\cite{zhang2023deep}. In this paper, we utilize Noisy Student~\cite{xie2020selftraining}. 

A teacher model uses labeled samples to train a supervised model, which is then applied to generate pseudo labels for unlabeled data. Following that, both the labeled and pseudo-labeled samples are used to re-train a noised student model. In our experiments, stochastic depth~\cite{huang2016deep} and heavy augmentation schemes inspired by RandAugment~\cite{cubuk2019randaugment} were applied to noise the student model. Soft pseudo labels were used to 
mitigate label imbalance and reduce the risk of propagating incorrect labels during training. 

Most of the external data did not have multi-view images for each study. Hence self-training was only applied to stage 1 pretraining of the convolutional neural networks. By default, we iterate the process 3 times.

\section{Experiment}

\subsection{Data}

\textbf{MIMIC-CXR} dataset~\cite{johnson2019mimiccxrjpg} is a large benchmark dataset for automated thorax disease classification. In the CXR-LT Shared Task, each CXR study in the dataset was labeled with 12 new rare disease findings following Holste \etal~\cite{Holste_2022}. The dataset contains 377110 chest X-rays, each labeled with at least one of 26 clinical findings (including a "No Finding" class). We use the split provided by the competition, with 264849 training images, 36769 validation images, and 75492 test images.

\textbf{CheXpert} dataset~\cite{irvin2019chexpert} is a large-scale Chest X-Ray dataset with 
223414 high-quality X-ray images in the train set. We use the CheXbert~\cite{smit2020chexbert} train labels with 14 class labels and 4 values: positive, negative, uncertain, and unmentioned. We only utilize the positive and negative labels for the overlapping classes with MIMIC-CXR. For uncertain, unmentioned labels or the 12 non-existing classes, we perform soft pseudo-labeling following Noisy Student~\cite{xie2020selftraining}. 

\textbf{NIH ChestX-ray14} dataset~\cite{Wang_2017} has 112,120 expert-annotated frontal-view X-rays
from 30805 unique patients with 14 disease labels. We only utilize train labels for the overlapping classes with MIMIC-CXR.
We utilize soft pseudo-labeling for the non-existing classes.

\textbf{VinDr-CXR} dataset~\cite{nguyen2022vindrcxr} consists of more than 15,000 chest X-ray scans that were retrospectively collected from two major hospitals in Vietnam. We use a modified version of the dataset from the VinBigData Chest X-ray Abnormalities Detection Competition.
The training set has 67914 images with 15 labels (including a "No Finding" class). Similarly, we only utilize the train labels for the overlapping classes with MIMIC-CXR. For non-existing classes, we perform soft pseudo-labeling. In contrast to MIMIC-CXR, the dataset has a Nodule/Mass class that treats the two symptoms as one class. When the Nodule/Mass class is assigned as positive, we compare the prediction results of the teacher model for the two classes and assign a hard positive label to the class with higher prediction probability and a soft label to the other. We assign negative labels to both classes when the class is assigned as negative.

\subsection{Implementation Details}
We leverage the ImageNet~\cite{imgnet,rw2019timm} pre-trained ConvNeXt~\cite{liu2022convnet} as our backbone. 
For ablation study purposes, we use ConvNeXt-T with images resized to 384 x 384. 
For multi-view model experiments and the final submission, we scale the model with ConvNeXt-S and use images resized to 1024 x 1024.

We use AdamW~\cite{loshchilov2019decoupled}, a learning rate of 3e-5, a cosine learning rate
schedule, no warmup, a batch size of 16, and a weight decay of 1e-2 for both stages. Test-time augmentation was used by averaging the predictions estimated from the original image and its horizontally flipped
image.

Training the final ConvNeXt-S model for submission takes 7 epochs, 50 hours for stage 1 and 6 epochs, 20 hours for stage 2, all on a single NVIDIA A6000 GPU. PyTorch was used for all implementations with public code available at: \url{https://github.com/dongkyuk/CXR-LT-public-solution}.

\begin{table*}
\begin{center}
\begin{tabular}{|l|l|c|c|c|c|c|}
\hline
\multirow{2}{*}{Backbone} & \multirow{2}{*}{Method}                         & \multicolumn{4}{|c|}{mAP} & AUC \\
\cline{3-7}
&   & total & head & medium & tail & total   \\
\hline\hline
\multirow{7}{*}{ConvNeXt-S-1024}    
& Single-view                       & 0.340 & 0.616 & 0.270 & 0.152 & 0.833 \\
& Multi-view Weighted Average (4:6) & 0.355 & 0.626& 0.284  & 0.170 &0.841         \\
& Multi-view Weighted Average (5:5) & 0.357 & 0.627& 0.288  & 0.172& 0.842   \\
& Multi-view Weighted Average (6:4) & 0.359 & 0.628& 0.296  & 0.167& 0.844        \\
& Multi-view Weighted Average (7:3) & 0.362 & 0.629& 0.306  & 0.167& 0.847        \\
& Multi-view Concat                 & 0.357 & 0.622& 0.293  & 0.173& 0.839        \\
& ChexFusion (Ours)                 & \textbf{0.372} & \textbf{0.630} & \textbf{0.312}  & \textbf{0.188} & \textbf{0.847}    \\
\hline
\end{tabular}
\end{center}
\caption{Performance comparison of ChexFusion and the baseline methods on the CXR-LT validation set.}
\end{table*}

\subsection{Evaluation Metrics}
We evaluate our method using metrics commonly used in multi-label classification tasks: Mean Average Precision (mAP) and Area Under the Receiver Operating Characteristic Curve (AUC-ROC). 

mAP is a popular evaluation metric that calculates the average precision for each class and then takes the mean over all classes. It measures the accuracy and ranking quality of the predicted probabilities for each class. On the other hand, AUC-ROC measures the trade-off between the true positive rate (sensitivity) and the false positive rate (1-specificity) across different classification thresholds.

In the CXR-LT shared task, the official evaluation metric is mAP, calculated separately for the validation and test sets. mAP is favored over AUC-ROC due to the strong imbalance in the dataset, which may lead to heavy inflation for AUC-ROC.

We split the 26 classes into three groups according to the number of training samples per class: 8 head classes, 10 medium classes and 8 tail classes. In the training set, the head classes have an average of 63918 samples per class, medium classes have an average of 10910 samples per class, and tail classes have an average of 1877 samples. For thorough evaluation, we report mAP results for each of the groups.

\subsection{Baselines}
We compare our method CheXFusion against several baselines to demonstrate its effectiveness.

\begin{itemize}
\item \textbf{Single-view:} We use the single-view model we trained during stage 1.

\item \textbf{Multi-view Weighted Average:} A simple yet powerful method for utilizing multiple views is to take a weighted average of the single-view model probability prediction results. We assign weight $w_f$ for frontal views such as Posterior Anterior (PA) and Anterior Posterior (AP) radiographs, and weight $w_l$ for Lateral and Left Lateral (LL) views, with $w_f + w_l = 1$. We report results with different values for the weights.

\item \textbf{Multi-view Concat:} A common baseline in multi-view models is to concatenate 1d features obtained from different views before being passed to a classification head. In this case, we train a single-view model with a GAP classification head, freeze the backbone, and then concatenate the features extracted from each view to train the final fully connected layer.
\end{itemize}

We use identical ConvNeXt-S backbone for all baselines with 1024 x 1024 input images. Test-time augmentation is used by default. 

\begin{table*}
\begin{center}
\begin{tabular}{|c|c|c|c|c|c|c|c|c|c|}
\hline
\multirow{2}{*}{wBCE} & \multirow{2}{*}{ASL~\cite{asl}} & \multirow{2}{*}{Ml-decoder~\cite{ridnik2021mldecoder}} & \multirow{2}{*}{Hard pseudo} & \multirow{2}{*}{Soft pseudo}          & \multicolumn{4}{|c|}{mAP} & AUC \\
\cline{6-10}
 &   & & & & total & head & medium & tail & total   \\
\hline\hline
  & & & & & 0.311 & 0.601 & 0.231 & 0.122 & 0.816 \\
 \centering \checkmark & & & & & 0.311 & 0.597 & 0.229 & 0.127 & 0.814 \\
  & \centering \checkmark & & & &  0.313 & 0.603 & 0.231 & 0.126 & 0.815 \\
 \centering \checkmark & \centering \checkmark & & & & 0.314 & 0.604 & 0.232 & 0.128 & 0.817 \\
 \centering \checkmark & \centering \checkmark & \centering \checkmark & & & 0.322 & 0.609 & 0.234 & 0.146 & 0.821 \\ 
 \centering \checkmark & \centering \checkmark & \centering \checkmark & \centering \checkmark & & 0.330 & 0.614 & 0.255 & 0.141 & 0.828 \\
 \centering \checkmark & \centering \checkmark & \centering \checkmark &  & \centering \checkmark & \textbf{0.336} & \textbf{0.612} & \textbf{0.270} & \textbf{0.143} & \textbf{0.832} \\
\hline
\end{tabular}
\end{center}
\caption{Ablation studies on the various components of our proposed method.}
\end{table*}

\subsection{Results}

Table 1 shows the performance comparison on the CXR-LT validation set. A simple weighted average of the single-view models achieves a significant performance increase compared to a single-view model. This proves the effectiveness of utilizing multiple views for improved classification. Another interesting observation is that higher values for the frontal view weight $w_f$ lead to better performance. This is likely due to the fact that frontal images provide a larger and clearer view of the chest, making it easier to detect and classify abnormalities. This shows that careful aggregation of multi-view features is crucial for optimal performance.

The concatenation model achieves similar mAP performance to the weighted average model but has a significantly lower AUC-ROC score. This suggests that the model may be more conservative in its predictions, leading to a higher specificity (lower false positive rate) but potentially sacrificing some sensitivity (lower true positive rate). In addition, the model shows lower scores for the head group with improved scores for the medium and tail groups.
These phenomena may be because the backbone was kept frozen during the training, and thus the decision boundaries specified by the jointly learned classifier were re-adjusted. The results are consistent with prior works from Kang \etal; on decoupling classifiers in long-tailed classification~\cite{kang2020decoupling}.

Our method, CheXFusion, outperforms all the baselines for both mAP and AUC-ROC. The total mAP achieved by CheXFusion on the validation set is 0.372, significantly higher than the best baseline result of 0.362. In addition, CheXFusion achieves higher mAP values for all three groups: head, medium, and tail. This indicates that our method effectively handles the imbalance between classes, including the rare and long-tailed classes. CheXFusion also scored 0.372 mAP with 0.850 AUC-ROC on the test set, winning first place in the CXR-LT competition. 

Overall, the results demonstrate that our proposed method, CheXFusion, effectively addresses the challenges of multi-view classification in chest X-ray analysis. It outperforms several baselines and achieves state-of-the-art performance on the CXR-LT dataset, indicating its potential for application in clinical settings.

\subsection{Ablation Study}

Table 2 presents the results of ablation studies on the various components of our proposed method. A ConvNeXt-T backbone with input images resized to 384 x 384 were used for experiment purposes.

First, we investigate the impact of different loss functions on the performance of the backbone model. We compare the use of binary cross-entropy (BCE) loss, weighted binary cross-entropy (wBCE) loss, asymmetric (ASL~\cite{asl}) loss, and a combination of ASL and Weighted BCE.
Using wBCE loss alone improves performance for the tail group due to the increased gradient update for tail group instances. However, it also reduces the gradient update of head and medium group instances, resulting in underfitting. Using ASL loss mitigates intra-class imbalances in all classes, increasing performance across all groups.
We can see that using both ASL and weighted BCE leads to the best performance, achieving a total mAP of 0.314 with the highest mAP values for all three groups. This indicates that the combination of these loss functions effectively handles both inter-class and intra-class imbalances in the long-tailed classification task, alleviating the head and medium group performance decrease in wBCE while keeping the tail group mAP improvement.

Next, we examine the effect of incorporating a transformer-based classification head, specifically Ml-decoder, in the backbone model. The results show that adding Ml-decoder improves the performance, with a total mAP of 0.322 compared to 0.314 without Ml-decoder. This suggests that the transformer-based classification head enhances the model's ability to capture local discriminative features adaptively for different classes.

Furthermore, we investigate the impact of self-training, specifically the Noisy Student method, on the performance of the backbone model. 
We explore the impact of using hard and soft pseudo-labeling for non-existing classes. Hard pseudo-labeling with a threshold of 0.5 improves overall performance but shows worse tail group mAP, possibly due to the increased label imbalance. Soft pseudo-labeling, on the other hand, leads to significantly better performance, with an mAP score of 0.336. This suggests that soft pseudo-labeling is more effective for long-tailed distributions. Overall, incorporating unlabeled data and pseudo-labeling can provide additional information and positive samples for training, leading to better generalization and performance.

The ablation study results confirm the effectiveness of each component in our proposed method. Combining weighted BCE loss, ASL, Ml-decoder, and self-training results in the highest performance, demonstrating each component's importance in addressing the challenges of multi-label long-tailed classification.

\begin{table}
\begin{center}
\begin{tabular}{|c|c|c|}
\hline
\multirow{2}{*}{Segment embedding} & \multirow{2}{*}{Shuffling} & mAP \\
\cline{3-3}
 & & total    \\
\hline\hline
& & 0.366 \\
\centering \checkmark & & 0.368 \\
\centering \checkmark & \centering \checkmark & \textbf{0.372} \\
\hline
\end{tabular}
\end{center}
\caption{Impact of segment embedding and shuffling on the performance of CheXFusion.}
\end{table}

\subsection{Effect of segment embedding and shuffling}
In this section, we analyze the effect of segment embedding and shuffling on the performance of our proposed method, CheXFusion.

Segment embedding is a learnable embedding added to each feature map to indicate which image it belongs to. This is similar to the segment embeddings used in natural language processing tasks, where different segments of a text sequence are encoded with different embeddings. In our case, each image is considered a segment, and adding segment embeddings helps the model distinguish between different images during the fusion process.

We experiment with two configurations: without segment embedding and with segment embedding. The results, shown in Table 3, demonstrate that incorporating segment embeddings leads to a performance increase. The total mAP improves from 0.366 without segment embeddings to 0.368 with segment embeddings. This indicates that segment embeddings provide appropriate context to the model.

Shuffling is another technique we employ to improve the performance of CheXFusion. By shuffling the feature maps along the first index, we ensure that the order of the feature maps is irrelevant to the task. 

Comparing the results of shuffling versus not shuffling, we observe that shuffling leads to the highest performance. The total mAP increases from 0.368 without shuffling to 0.372 with shuffling. This indicates that shuffling is crucial in preventing the model from overfitting to the training image order distribution.

Overall, incorporating segment embedding and shuffling significantly improves the performance of CheXFusion. These techniques enhance the model's ability to capture and integrate features from multiple views and remove possible order bias during training.

\subsection{Future Work}
Our proposed method, CheXFusion, demonstrates strong performance in the multi-view classification of chest X-ray images. However, there are potential areas for future improvement:

    \textbf{Joint training}: 
Our current approach freezes the pre-trained backbone during the training of the fusion module. While this allows faster training and focuses the learning on the fusion module, there is still room for exploration in terms of joint training. By jointly training the backbone and fusion module, we can potentially achieve better integration of multi-view features and optimize the overall performance of the model. 
    
    \textbf{View-specific backbone}: 
Our current approach uses a single backbone for all views. While this reduces model complexity, it may not fully capture the unique characteristics and features present in each view. Future work can explore view-specific backbones with a Mixture of experts (MoE)~\cite{shazeer2017outrageously}, allowing the model to learn and leverage view-specific information more effectively with minimal model complexity. This could lead to improved performance and better utilization of multi-view features.

    \textbf{Interpretability and explainability}: 
Our current method, like many deep learning models, lacks explicit interpretability. Chest X-ray analysis is a critical domain where interpretability is highly valuable for clinicians to understand and trust the model's predictions. Future work can focus on integrating interpretability techniques into our method to provide insights into the model's decision-making process. The attention mechanisms in our work allow using visualization methods and saliency maps to generate intuitive visual explanations. It also allows easy integration with language models, where we can explicitly output the model's reasoning. 

\subsection{Conclusion}
In this paper, we proposed CheXFusion, a module for multi-view classification of chest X-ray images. Our method consists of two stages: backbone pre-training and transformer fusion model. In the backbone pr-etraining stage, we trained a single-view convolutional neural network backbone with Ml-Decoder, a transformer-based classification head. In the fusion stage, we trained a transformer-based fusion module that effectively integrates the features extracted from multiple views. Furthermore, we explore the effectiveness of data balancing techniques and self-training strategies in multi-label long-tailed classification through extensive experiments. CheXFusion shows top-1-ranking performance in the CXR-LT competition. 
\section{Acknowledgments}
We thank Alex Xie for the naming of the model and the valuable discussions.

{\small
\bibliographystyle{ieee_fullname}
\bibliography{egbib}

\begin{thebibliography}{10}\itemsep=-1pt

\bibitem{asl}
Emanuel Ben-Baruch, Tal Ridnik, Nadav Zamir, Asaf Noy, Itamar Friedman, Matan
  Protter, and Lihi Zelnik-Manor.
\newblock Asymmetric loss for multi-label classification, 2021.

\bibitem{carion2020endtoend}
Nicolas Carion, Francisco Massa, Gabriel Synnaeve, Nicolas Usunier, Alexander
  Kirillov, and Sergey Zagoruyko.
\newblock End-to-end object detection with transformers, 2020.

\bibitem{chen2019learning}
Tianshui Chen, Muxin Xu, Xiaolu Hui, Hefeng Wu, and Liang Lin.
\newblock Learning semantic-specific graph representation for multi-label image
  recognition, 2019.

\bibitem{cubuk2019randaugment}
Ekin~D. Cubuk, Barret Zoph, Jonathon Shlens, and Quoc~V. Le.
\newblock Randaugment: Practical automated data augmentation with a reduced
  search space, 2019.

\bibitem{devlin2019bert}
Jacob Devlin, Ming-Wei Chang, Kenton Lee, and Kristina Toutanova.
\newblock Bert: Pre-training of deep bidirectional transformers for language
  understanding, 2019.

\bibitem{dosovitskiy2021image}
Alexey Dosovitskiy, Lucas Beyer, Alexander Kolesnikov, Dirk Weissenborn,
  Xiaohua Zhai, Thomas Unterthiner, Mostafa Dehghani, Matthias Minderer, Georg
  Heigold, Sylvain Gelly, Jakob Uszkoreit, and Neil Houlsby.
\newblock An image is worth 16x16 words: Transformers for image recognition at
  scale, 2021.

\bibitem{FEIGIN20101560}
David~S. Feigin.
\newblock Lateral chest radiograph: A systematic approach.
\newblock {\em Academic Radiology}, 17(12):1560--1566, 2010.

\bibitem{Holste_2022}
Gregory Holste, Song Wang, Ziyu Jiang, Thomas~C. Shen, George Shih, Ronald~M.
  Summers, Yifan Peng, and Zhangyang Wang.
\newblock Long-tailed classification of~thorax diseases on~chest x-ray: A new
  benchmark study.
\newblock In {\em Lecture Notes in Computer Science}, pages 22--32. Springer
  Nature Switzerland, 2022.

\bibitem{huang2016deep}
Gao Huang, Yu Sun, Zhuang Liu, Daniel Sedra, and Kilian Weinberger.
\newblock Deep networks with stochastic depth, 2016.

\bibitem{irvin2019chexpert}
Jeremy Irvin, Pranav Rajpurkar, Michael Ko, Yifan Yu, Silviana Ciurea-Ilcus,
  Chris Chute, Henrik Marklund, Behzad Haghgoo, Robyn Ball, Katie Shpanskaya,
  Jayne Seekins, David~A. Mong, Safwan~S. Halabi, Jesse~K. Sandberg, Ricky
  Jones, David~B. Larson, Curtis~P. Langlotz, Bhavik~N. Patel, Matthew~P.
  Lungren, and Andrew~Y. Ng.
\newblock Chexpert: A large chest radiograph dataset with uncertainty labels
  and expert comparison, 2019.

\bibitem{johnson2019mimiccxrjpg}
Alistair E.~W. Johnson, Tom~J. Pollard, Nathaniel~R. Greenbaum, Matthew~P.
  Lungren, Chih ying Deng, Yifan Peng, Zhiyong Lu, Roger~G. Mark, Seth~J.
  Berkowitz, and Steven Horng.
\newblock Mimic-cxr-jpg, a large publicly available database of labeled chest
  radiographs, 2019.

\bibitem{kang2020decoupling}
Bingyi Kang, Saining Xie, Marcus Rohrbach, Zhicheng Yan, Albert Gordo, Jiashi
  Feng, and Yannis Kalantidis.
\newblock Decoupling representation and classifier for long-tailed recognition,
  2020.

\bibitem{10020356}
Mohammad Kohankhaki, Ahmad Ayad, Mahdi Barhoush, Bastian Leibe, and Anke
  Schmeink.
\newblock Radiopaths: Deep multimodal analysis on chest radiographs.
\newblock In {\em 2022 IEEE International Conference on Big Data (Big Data)},
  pages 3613--3621, 2022.

\bibitem{imgnet}
Alex Krizhevsky, Ilya Sutskever, and Geoffrey Hinton.
\newblock Imagenet classification with deep convolutional neural networks.
\newblock {\em Neural Information Processing Systems}, 25, 01 2012.

\bibitem{726791}
Y. Lecun, L. Bottou, Y. Bengio, and P. Haffner.
\newblock Gradient-based learning applied to document recognition.
\newblock {\em Proceedings of the IEEE}, 86(11):2278--2324, 1998.

\bibitem{liu2021query2label}
Shilong Liu, Lei Zhang, Xiao Yang, Hang Su, and Jun Zhu.
\newblock Query2label: A simple transformer way to multi-label classification,
  2021.

\bibitem{liu2022convnet}
Zhuang Liu, Hanzi Mao, Chao-Yuan Wu, Christoph Feichtenhofer, Trevor Darrell,
  and Saining Xie.
\newblock A convnet for the 2020s, 2022.

\bibitem{loshchilov2019decoupled}
Ilya Loshchilov and Frank Hutter.
\newblock Decoupled weight decay regularization, 2019.

\bibitem{nguyen2022vindrcxr}
Ha~Q. Nguyen, Khanh Lam, Linh~T. Le, Hieu~H. Pham, Dat~Q. Tran, Dung~B. Nguyen,
  Dung~D. Le, Chi~M. Pham, Hang T.~T. Tong, Diep~H. Dinh, Cuong~D. Do, Luu~T.
  Doan, Cuong~N. Nguyen, Binh~T. Nguyen, Que~V. Nguyen, Au~D. Hoang, Hien~N.
  Phan, Anh~T. Nguyen, Phuong~H. Ho, Dat~T. Ngo, Nghia~T. Nguyen, Nhan~T.
  Nguyen, Minh Dao, and Van Vu.
\newblock Vindr-cxr: An open dataset of chest x-rays with radiologist's
  annotations, 2022.

\bibitem{raffel2020exploring}
Colin Raffel, Noam Shazeer, Adam Roberts, Katherine Lee, Sharan Narang, Michael
  Matena, Yanqi Zhou, Wei Li, and Peter~J. Liu.
\newblock Exploring the limits of transfer learning with a unified text-to-text
  transformer, 2020.

\bibitem{RAOOF2012545}
Suhail Raoof, David Feigin, Arthur Sung, Sabiha Raoof, Lavanya Irugulpati, and
  Edward~C. Rosenow.
\newblock Interpretation of plain chest roentgenogram.
\newblock {\em Chest}, 141(2):545--558, 2012.

\bibitem{ClassifierChains}
Jesse Read, Bernhard Pfahringer, Geoffrey Holmes, and Eibe Frank.
\newblock Classifier chains for multi-label classification.
\newblock volume~85, pages 254--269, 08 2009.

\bibitem{ridnik2021mldecoder}
Tal Ridnik, Gilad Sharir, Avi Ben-Cohen, Emanuel Ben-Baruch, and Asaf Noy.
\newblock Ml-decoder: Scalable and versatile classification head, 2021.

\bibitem{rubin2018large}
Jonathan Rubin, Deepan Sanghavi, Claire Zhao, Kathy Lee, Ashequl Qadir, and
  Minnan Xu-Wilson.
\newblock Large scale automated reading of frontal and lateral chest x-rays
  using dual convolutional neural networks, 2018.

\bibitem{shazeer2017outrageously}
Noam Shazeer, Azalia Mirhoseini, Krzysztof Maziarz, Andy Davis, Quoc Le,
  Geoffrey Hinton, and Jeff Dean.
\newblock Outrageously large neural networks: The sparsely-gated
  mixture-of-experts layer, 2017.

\bibitem{smit2020chexbert}
Akshay Smit, Saahil Jain, Pranav Rajpurkar, Anuj Pareek, Andrew~Y. Ng, and
  Matthew~P. Lungren.
\newblock Chexbert: Combining automatic labelers and expert annotations for
  accurate radiology report labeling using bert, 2020.

\bibitem{Multi-Label}
Grigorios Tsoumakas and Ioannis Katakis.
\newblock Multi-label classification: An overview.
\newblock {\em International Journal of Data Warehousing and Mining}, 3:1--13,
  09 2009.

\bibitem{Tsoumakas}
Grigorios Tsoumakas, Ioannis Katakis, and I. Vlahavas.
\newblock {\em Mining Multi-label Data}, pages 667--685.
\newblock 07 2010.

\bibitem{vaswani2017attention}
Ashish Vaswani, Noam Shazeer, Niki Parmar, Jakob Uszkoreit, Llion Jones,
  Aidan~N. Gomez, Lukasz Kaiser, and Illia Polosukhin.
\newblock Attention is all you need, 2017.

\bibitem{Wang_2017}
Xiaosong Wang, Yifan Peng, Le Lu, Zhiyong Lu, Mohammadhadi Bagheri, and
  Ronald~M. Summers.
\newblock {ChestX}-ray8: Hospital-scale chest x-ray database and benchmarks on
  weakly-supervised classification and localization of common thorax diseases.
\newblock In {\em 2017 {IEEE} Conference on Computer Vision and Pattern
  Recognition ({CVPR})}. {IEEE}, jul 2017.

\bibitem{wang2017multilabel}
Zhouxia Wang, Tianshui Chen, Guanbin Li, Ruijia Xu, and Liang Lin.
\newblock Multi-label image recognition by recurrently discovering attentional
  regions, 2017.

\bibitem{rw2019timm}
Ross Wightman.
\newblock Pytorch image models.
\newblock \url{https://github.com/rwightman/pytorch-image-models}, 2019.

\bibitem{dbloss}
Tong Wu, Qingqiu Huang, Ziwei Liu, Yu Wang, and Dahua Lin.
\newblock Distribution-balanced loss for multi-label classification in
  long-tailed datasets, 2021.

\bibitem{xie2020selftraining}
Qizhe Xie, Minh-Thang Luong, Eduard Hovy, and Quoc~V. Le.
\newblock Self-training with noisy student improves imagenet classification,
  2020.

\bibitem{ye2020attentiondriven}
Jin Ye, Junjun He, Xiaojiang Peng, Wenhao Wu, and Yu Qiao.
\newblock Attention-driven dynamic graph convolutional network for multi-label
  image recognition, 2020.

\bibitem{ML-KNN}
Min-Ling Zhang and Zhi-Hua Zhou.
\newblock Ml-knn: A lazy learning approach to multi-label learning.
\newblock {\em Pattern Recognition}, 40(7):2038--2048, 2007.

\bibitem{zhang2023deep}
Yifan Zhang, Bingyi Kang, Bryan Hooi, Shuicheng Yan, and Jiashi Feng.
\newblock Deep long-tailed learning: A survey, 2023.

\bibitem{Zhou_2021}
S.~Kevin Zhou, Hayit Greenspan, Christos Davatzikos, James~S. Duncan, Bram~Van
  Ginneken, Anant Madabhushi, Jerry~L. Prince, Daniel Rueckert, and Ronald~M.
  Summers.
\newblock A review of deep learning in medical imaging: Imaging traits,
  technology trends, case studies with progress highlights, and future
  promises.
\newblock {\em Proceedings of the {IEEE}}, 109(5):820--838, may 2021.

\bibitem{mvcnet}
Xiongfeng Zhu and Qianjin Feng.
\newblock Mvc-net: Multi-view chest radiograph classification network with deep
  fusion.
\newblock pages 554--558, 04 2021.

\end{thebibliography}
}

\end{document}